\documentclass[runningheads]{llncs}
\usepackage{amsmath}
\usepackage{accessibility}
\usepackage{booktabs}
\usepackage{tabularx}
\usepackage[T1]{fontenc}
\usepackage{graphicx}
\begin{document}
\title{Multi-Constrained Evolutionary Molecular Design Framework: An Interpretable Drug Design Method Combining Rule-Based Evolution and Molecular Crossover}
%
%
%

\author{Shanxian Lin\inst{1}\orcidID{0009-0001-4855-8252} \and Wei Xia\inst{2} \and Yuichi Nagata\inst{1}\orcidID{0000-0001-7582-2699}$^*$ \and Haichuan Yang\inst{1}\orcidID{0000-0001-7100-7945}$^*$\thanks{$^*$Corresponding authors}}

\authorrunning{S. Lin et al.}

\institute{Graduate School of Technology, Industrial and Social Sciences, Tokushima University, Tokushima 770-8506, Japan\\
\email{utigooselsx@gmail.com, nagata@is.tokushima-u.ac.jp, you.kaisen@tokushima-u.ac.jp} \and
College of Chemistry and Chemical Engineering, China University of Petroleum (East China), Qingdao 266580, Shandong, China\\
\email{xiawei@upc.edu.cn}}

\maketitle              
\begin{abstract}
This study proposes MCEMOL (Multi-Constrained Evolutionary Molecular Design Framework), a molecular optimization approach integrating rule-based evolution with molecular crossover. MCEMOL employs dual-layer evolution: optimizing transformation rules at rule level while applying crossover and mutation to molecular structures. Unlike deep learning methods requiring large datasets and extensive training, our algorithm evolves efficiently from minimal starting molecules with low computational overhead.
The framework incorporates message-passing neural networks and comprehensive chemical constraints, ensuring efficient and interpretable molecular design. Experimental results demonstrate that MCEMOL provides transparent design pathways through its evolutionary mechanism while generating valid, diverse, target-compliant molecules. The framework achieves 100\% molecular validity with high structural diversity and excellent drug-likeness compliance, showing strong performance in symmetry constraints, pharmacophore optimization, and stereochemical integrity.
Unlike black-box methods, MCEMOL delivers dual value: interpretable transformation rules researchers can understand and trust, alongside high-quality molecular libraries for practical applications. This establishes a paradigm where interpretable AI-driven drug design and effective molecular generation are achieved simultaneously, bridging the gap between computational innovation and practical drug discovery needs.

\keywords{Evolutionary algorithm  \and Drug design \and Molecular optimization \and Interpretable artificial intelligence \and Molecular crossover.}
\end{abstract}
\section{Introduction}

Molecular design traditionally relies on expert intuition and trial-and-error, making the discovery process slow and costly. Drug development can take 10-15 years, with costs exceeding \$1\,billion and success rates below 10\% \cite{Paul2010}. Computer-aided molecular design improves this process by prioritizing candidates in silico, cutting experimental costs, and enhancing target specificity and safety.

Recent machine learning (ML) methods have advanced molecular generation \cite{doi:10.1126/science.aat2663,C9ME00039A}, yet they face three main challenges: (i) ensuring chemical validity and synthetic feasibility, (ii) avoiding loss of structural diversity, and (iii) providing interpretable designs \cite{Gao2020,Maziarka2020}.

Conventional methods such as variational autoencoders (VAEs) \cite{Gomez-Bombarelli2018} and generative adversarial networks (GANs) \cite{2018arXiv180511973D} have shown promise but often generate invalid or non-synthesizable molecules. Reinforcement learning (RL) methods optimize specific properties, but lack interpretability and may overfit \cite{Zhou2019,NEURIPS2018_d60678e8}. Rule-based systems preserve validity but are limited by fixed transformation sets \cite{Polishchuk2020,Segler2018}.

We propose MCEMOL, a Multi-Constrained Evolutionary Molecular Design Framework, that combines rule-based evolution and molecular-level crossover. MCEMOL uses a dual-layer evolutionary approach: it evolves both molecular structures and transformation rules, enhancing design flexibility while ensuring chemical validity. Crucially, unlike deep learning methods requiring extensive computational resources and large datasets, MCEMOL operates efficiently with minimal computational overhead and can evolve from just a few starting molecules. By integrating a message-passing neural network (ATLAS-DMPNN) and a multidimensional chemical constraint system, MCEMOL optimizes molecules with diverse properties, including synthetic feasibility, symmetry, and pharmacophore integrity. Unlike "black-box" methods, MCEMOL provides dual outputs: an interpretable design pathway through discovered transformation rules, and a high-quality molecular library with 100\% validity and excellent drug-likeness compliance. This dual capability ensures that researchers gain both mechanistic insights and practical molecular candidates for further development.
Our contributions include: (1) introducing a constrained evolutionary framework that generates multi-property-targeted molecules while ensuring chemical validity, (2) unifying rule-level learning with molecule-level evolution to maintain diversity and synthesizability, (3) discovering empirically validated rules governing reactions of specific drug molecules, (4) delivering a ready-to-use molecular library alongside the discovered transformation rules, demonstrating that interpretability and generation quality are not mutually exclusive, and (5) achieving computational efficiency that enables practical deployment without requiring large-scale datasets or extensive training periods.

\section{Literature Review}

Machine learning approaches to molecular design have evolved through multiple paradigms. Deep generative models, particularly VAEs and GANs, pioneered the use of continuous latent spaces for molecular generation. Gómez-Bombarelli et al. \cite{Gomez-Bombarelli2018} encoded SMILES strings into latent representations for gradient-based optimization, while Jin et al. \cite{pmlr-v80-jin18a} extended this to molecular graphs. Flow-based models \cite{2020arXiv200109382S} introduced invertible transformations between data and latent spaces. However, these methods frequently produce invalid or unsynthesizable molecules.

Reinforcement learning treats molecular generation as sequential decision-making with reward-guided policies. Zhou et al. \cite{Zhou2019} applied policy gradient methods for property optimization, while You et al. \cite{NEURIPS2018_d60678e8} developed graph-based policies for multi-constraint satisfaction. Despite effective property control, RL methods suffer from poor interpretability and policy collapse that reduces structural diversity. Graph neural networks \cite{pmlr-v70-gilmer17a,NIPS2015_f9be311e} provide essential property prediction capabilities to guide these optimization processes.

Rule-based systems leverage chemical knowledge to ensure validity and synthesizability. CReM \cite{Polishchuk2020} and synthesis-planning frameworks \cite{Segler2018} apply predefined transformation rules, but static rule sets limit novel chemotype discovery. Evolutionary algorithms address this limitation through selection, crossover, and mutation operations \cite{C8SC05372C,https://doi.org/10.1002/minf.201600044}, though maintaining chemical validity remains challenging.

Evolutionary Rule Learning (ERL) systems combine evolutionary search with rule-based interpretability. Systems like XCS \cite{https://doi.org/10.1155/2009/736398} and Genetic Fuzzy Systems \cite{943725} adapt rule sets while managing uncertainty. Chemistry-aware crossover operators, including fragment recombination \cite{Fechner2006} and BRICS decomposition \cite{Degen2008chem}, preserve key substructures during evolution.

Molecular design requires balancing structural validity (bonding, valency, stereochemistry) \cite{Halgren1996MerckMF,Brooks2011}, synthesizability (SA-Score) \cite{Ertl2009}, and pharmacological properties. Drug-likeness filters \cite{LIPINSKI19973,Veber2002} and pharmacophore patterns \cite{Wolber2005} guide optimization. Effective design navigates trade-offs between potency, ADMET properties, and complexity \cite{NICOLAOU2013e427} while maintaining interpretability and chemical validity.

\section{Framework Components}
\label{sec3}
To address the aforementioned challenges in molecular design, we propose the Multi-dimensional Constrained Evolutionary Molecular Design Framework (MCEMOL). This framework targets chemical validity, property control, and interpretability through a dual-layer evolutionary approach that simultaneously optimizes transformation rules and directly edits molecular structures. Distinct from data-intensive deep learning models, MCEMOL requires minimal computational resources, capable of effectively evolving from a limited number of starting molecules without extensive training. This approach ensures traceability via interpretable rule representations while generating a diverse library of valid molecules. By employing a multi-strategy molecular crossover engine, MCEMOL enhances structural diversity and innovation, effectively balancing interpretability with generation performance. This section outlines the framework's foundational components, followed by their integration and extension into our proposed system.

\subsection{Rule Representation System.}
\label{sec3-1}
Following established evolutionary rule learning principles \cite{Bacardit2009}, we adopt a three-component rule representation, providing an interpretable foundation for our framework and enabling chemists to understand and trust the molecular transformations during design.

\begin{equation}
    R = (C, T, w_r)
\end{equation}
where the condition set \( C \) defines the applicability criteria of the rule, the mutation set \( T \) specifies the operations to be executed, and the rule weight \( w_r \) serves as a multiplicative factor in fitness evaluation, influencing the rule's selection probability during evolutionary operations.

The condition set \( C \) determines rule applicability via six categories: structural (e.g., aromatic rings), physicochemical (e.g., MW, LogP), pharmacological, stability (excluding unstable moieties), drug-likeness (Lipinski/Veber rules), and advanced (synthetic/metabolic assessment). Each condition returns a boolean value to ensure precise transformation control.

The transformation set \( T \) includes seven mutation types with 36 functions, such as group addition (e.g., methylation), atom substitution (e.g., carbon to nitrogen), group modification, and ring manipulation (e.g., adding a benzene ring). For example, phenol can be mutated into cyclohexanol, 4-hydroxypyridine, anisole, or benzo[cyclo]hexanol.

The application of a rule follows a defined process: when a molecule \( m \in M \) satisfies all conditions in \( C \), the mutations in \( T \) are applied sequentially from left to right; otherwise, the function returns null:
\begin{equation}
    \label{eq:apply}
    \operatorname{apply}(R, m) =
    \begin{cases}
        t_1 \circ t_2 \circ \dots \circ t_k (m), & \text{if } \forall c_j \in C,\; c_j(m)=\mathrm{true}, \\
        \mathrm{null}, & \text{otherwise},
    \end{cases}
\end{equation}
where \(M\) denotes the current molecule set (population), and \(m\) is a single molecule with \(m\in M\); \(C=\{c_j\}\) is the set of conditions; and \(T=\langle t_1,\ldots,t_k\rangle\) is the ordered list of mutations, where \(k\) is the number of mutations. The composition \(t_1 \circ t_2 \circ \cdots \circ t_k\) is \emph{defined} to be evaluated left-to-right, meaning that molecule \(m\) undergoes \(k\) successive transformations \(t_1,\ldots,t_k\); if any transformation yields \(\mathrm{null}\), the entire composition returns \(\mathrm{null}\).

This representation is both machine-processable and intuitive for chemists, forming a key foundation for the interpretability of the MCEMOL framework \cite{SWAN2022393}.

\subsection{Multidimensional Chemical Constraint System and Property Prediction.}
\label{sec3-2}
We employ a comprehensive constraint system with twelve complementary constraints \cite{Ertl2009,Veber2002}, ensuring that generated molecules possess desired properties and meet practical drug development requirements. These constraints filter out theoretically interesting but impractical candidates while enabling accurate prediction of complex pharmacological properties that cannot be derived from simple structural rules.

The constraints are categorized as \emph{basic} and \emph{advanced} criteria. Basic criteria enforce validity and core drug-like behavior: (i) fraction of chemically valid molecules, (ii) point-group symmetry \cite{Halgren1996MerckMF}, (iii) property prediction via message-passing neural networks, (iv) fingerprint-based structural diversity, and (v) key physicochemical traits like logP and TPSA \cite{LIPINSKI19973}.

Advanced criteria focus on developability and function: (i) synthetic feasibility \cite{Ertl2009}, (ii) reactivity/stability screening, (iii) pharmacophore compliance \cite{Wolber2005}, (iv) conformational flexibility, (v) scaffold preservation, (vi) stereochemical integrity \cite{Brooks2011}, and (vii) extended drug-likeness from medicinal-chemistry guidelines \cite{Veber2002}.

For property prediction, we employ ATLAS-DMPNN exclusively for complex pharmacological endpoints, whereas fundamental physicochemical properties (molecular weight, logP, TPSA) are calculated directly via established algorithms. Given that ADMET endpoints require machine learning to capture non-linear structure-activity relationships, ATLAS-DMPNN addresses this through directed message passing with attention-guided aggregation, topology-aware encodings, and structural motif mining \cite{Lin2025ATLAS-DMPNN}. The model processes molecular graphs containing atom (type, charge, hybridization, aromaticity) and bond features (order, ring membership, stereochemistry) to output predicted property values. We adopt the original architecture (hidden dimension 300, three message-passing layers, dropout 0.2), training from scratch on task-specific datasets. These features enable the identification of metabolic soft spots, toxicophores, and protein-binding sites—critical patterns for clinical success often missed by standard MPNNs due to limited context modeling.

Each candidate molecule is scored based on these constraints and predicted properties, where each property \(S_u(m)\) is normalized and aggregated:
\begin{equation}
    \label{eq:fm}
    F(m) = \sum_{i=1}^{n} w_u \cdot S_u(m)
\end{equation}
where \( w_u \) is the weight for property \( u \), and \( S_u(m) \in [0,1] \) is its normalized score combining both directly calculated values and ATLAS-DMPNN predictions. Weights are task-specific, prioritizing design goals \cite{NICOLAOU2013e427}. 

For central nervous system (CNS) drugs, stringent constraints are required for blood-brain barrier (BBB) penetration: LogP values of 2.8 (range 2-3), molecular weight around 305 Da, TPSA of 44.8 \AA\ (optimal range <90 \AA), \(\leq\)2 hydrogen bond donors, and moderate basicity (pKa ~8.4) \cite{Wager2010}. The CNS MPO score integrates these properties, with scores \(\geq\)4.0 indicating favorable BBB penetration \cite{Wager2016}. This integrated approach ensures computational resources are prioritized on properties requiring machine learning, rather than those accessible via deterministic calculations, while maintaining comprehensive molecular evaluation throughout the evolutionary process.

\subsection{Molecular Crossover Strategies and Connection Point Selection.}
\label{sec3-3}
MCEMOL employs seven complementary crossover strategies with intelligent connection-point selection to ensure that products are chemically valid and structurally novel. These strategies generate diverse molecular offspring while maintaining chemical feasibility, essential for exploring broad chemical space without producing invalid structures.

The seven crossover strategies are:
(i) Fragment-based crossover, which retains the central scaffold of one parent while replacing peripheral functional groups from another, preserving the core pharmacophore structure \cite{Lewell1998}.
(ii) Scaffold-based crossover, which replaces the central scaffold while preserving peripheral side chains, enabling exploration of novel chemotypes that maintain similar binding interactions \cite{Bemis1996}.
(iii) BRICS crossover, which recombines at synthetically reasonable cleavage points to enhance accessibility \cite{Degen2008chem}.
(iv) Substructure exchange, which swaps rings or functional groups between parent molecules \cite{Ertl2003}.
(v) Pharmacophore-guided recombination, which retains key pharmacophoric features during crossover \cite{Wolber2005}.
(vi) Multi-parent recombination, which selects complementary fragments from three or more parents through tournament selection based on fragment fitness scores, then assembles them into a single offspring \cite{Nicolaou2009}.
(vii) Reaction-template-based crossover, which uses reaction templates to improve synthetic feasibility \cite{Coley2017}.

For connection point selection, we use four heuristic strategies with weighted probabilistic selection, where strategy weights are updated based on historical success rates:
(i) Best match, which identifies atoms with similar chemical environments using Morgan fingerprints \cite{Rogers2010}.
(ii) Reactive groups, which utilize predefined reactive group pairs for feasible bond formation \cite{Hartenfeller2011}.
(iii) Minimize strain, which prioritizes sterically accessible atoms based on their degree and hydrogen count \cite{Riniker2015}.
(iv) Preserve features, which avoids disrupting identified pharmacophoric elements \cite{Dixon2006}.


\section{Methods}

\begin{figure}[t]
\begin{center}
\includegraphics[width=\textwidth]{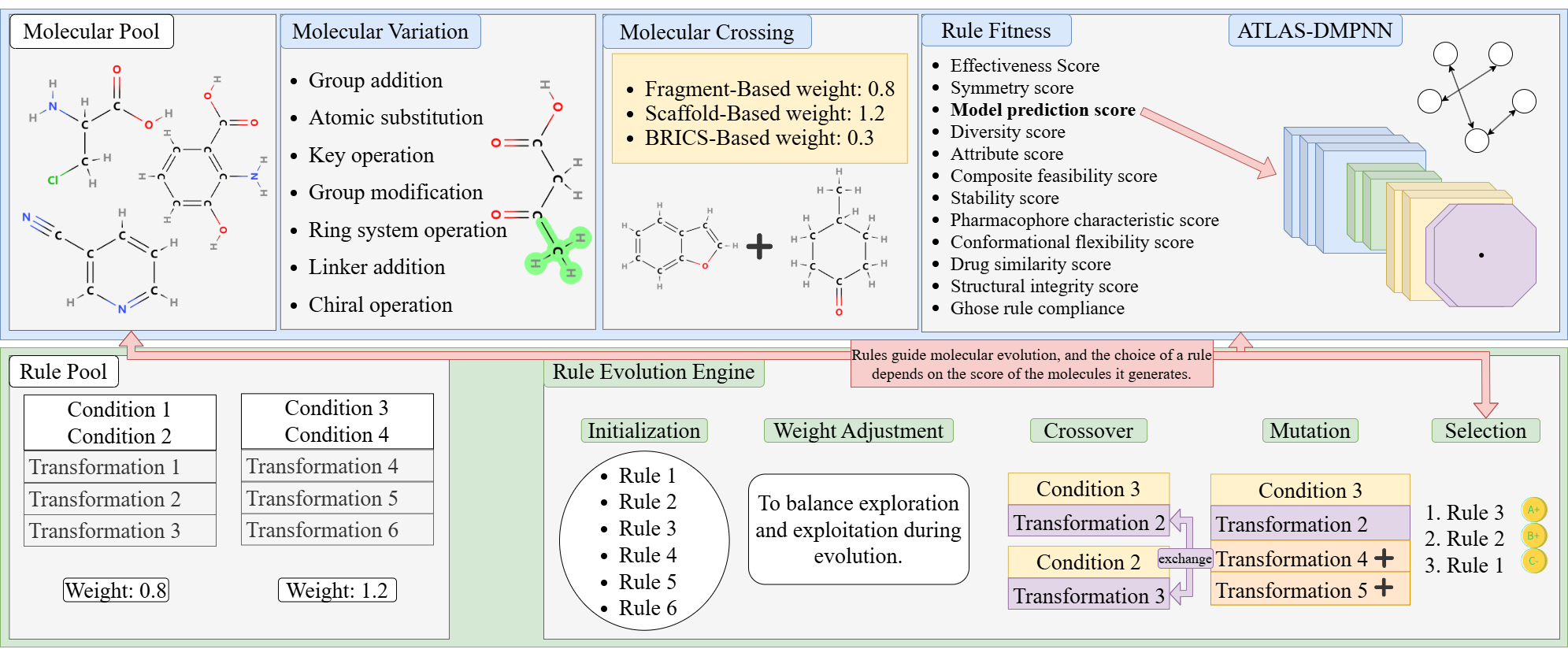}
\end{center}
\caption{MCEMOL's overall architecture.}
\label{all}
\end{figure}

As shown in Fig.~\ref{all}, MCEMOL comprises two integrated components: the molecular evolution engine (top, labeled in blue) and the rule evolution engine (bottom, labeled in green). The following subsections detail each component.

\subsection{Rule Evolution Engine}
\label{sec:rule-evo}

Building on the rule representation in Section~\ref{sec3-1}, we develop an adaptive genetic algorithm to continuously evolve the rule set \( R \).  To mitigate the risk of premature convergence caused by local optima in high-dimensional combinatorial spaces, MCEMOL dynamically optimizes the rule set. This ensures effective exploitation within a finite scope, thereby enhancing molecular design performance.

The evolution process consists of four key steps:

\textbf{Initialization}: The initial rule population is generated with diverse conditions and transformations to ensure broad coverage of chemical space.

\textbf{Weight Adjustment}: To balance exploration and exploitation during evolution, we adopt a weight update mechanism inspired by reinforcement learning's policy gradient methods. Rule weights are updated based on performance feedback as follows:
\begin{equation}
    w_r = w_r + \alpha \cdot (r - b) + \eta
\end{equation}
where \( w_r \) is the rule weight (distinct from the property weights \( w_u \) in Eq.~\eqref{eq:fm}), \( \alpha \) is the learning rate, \( r \) is the reward based on rule performance, \( b \) is the baseline to reduce variance, and \( \eta \) is exploration noise to prevent premature convergence.

\textbf{Evolutionary Operations}: The rule set undergoes iterative refinement through selection, crossover, and mutation.

First, crossover generates offspring rules by combining parent rules. Since a rule $R = (C, T, w_r)$ consists of three components, the conditions and transformations are crossed over independently:
\begin{equation}
    R^{\prime i} = 
    \begin{cases}
        (C^{\prime}, T^{\prime}, \bar{w}_r), & \text{if } \rho < p_c \\
        R^i_{\text{p1}}, & \text{otherwise}
    \end{cases}
\end{equation}
where
\begin{equation}
    C^{\prime} = C_{\text{p1}}[:\xi_C] \oplus C_{\text{p2}}[\xi_C:], \quad
    T^{\prime} = T_{\text{p1}}[:\xi_T] \oplus T_{\text{p2}}[\xi_T:], \quad
    \bar{w}_r = \frac{w_{r,\text{p1}} + w_{r,\text{p2}}}{2}
\end{equation}
Here, p1 and p2 denote parent1 and parent2 respectively, $C[:\xi]$ represents the first $\xi$ elements of list $C$, $C[\xi:]$ represents elements from position $\xi$ onward, $\oplus$ denotes list concatenation, and $\xi_C$, $\xi_T$ are random crossover points within the respective list lengths. Crossover occurs with probability $p_c$ when $\rho \sim \mathcal{U}(0,1)$ satisfies $\rho < p_c$.

Second, mutation introduces variations to maintain diversity:
\begin{equation}
    R^{\prime\prime i} = 
    \begin{cases}
        \Delta_\sigma(R^{\prime i}), & \text{if } \rho < p_m \\
        R^{\prime i}, & \text{otherwise}
    \end{cases}
\end{equation}
where $\rho \sim \mathcal{U}(0,1)$, $p_m$ is the mutation probability, and $\Delta_\sigma$ represents a stochastic perturbation operator. Mutation is applied independently to both conditions $C$ and transformations $T$. For conditions, strategy $\sigma \in \{\text{add}, \text{remove}, \text{replace}\}$ is selected probabilistically:
(i) Add: appends a random condition from the available pool.
(ii) Remove: deletes an existing condition (maintaining at least one).
(iii) Replace: $\Delta_\sigma$ substitutes an existing element with a new one.
(iv) Reorder: $\Delta_\sigma$ applies a random permutation exclusively to the transformation sequence $T$.

Finally, the evolutionary update process determines the next generation of rules through fitness-based selection. The population-level fitness is computed as:
\begin{equation}
    \overline{F}(M) = \frac{1}{|M|}\sum_{m \in M} F(m)
    \label{eq:fitness_function}
\end{equation}
where $F(m)$ is the molecular fitness function from Section~\ref{sec3-2} (Eq.~\eqref{eq:fm}), and $M$ denotes the molecular population. The selection mechanism then updates the rule population:
\begin{equation}
    R^{i+1} = 
    \begin{cases}
        R^{\prime\prime i}, & \text{if } \overline{F}(M^{i+1}) > \overline{F}(M^i) \\
        R^i, & \text{otherwise}
    \end{cases}
    \label{eq:rule_update}
\end{equation}
where $R^{\prime\prime i}$ represents the candidate rule population after crossover and mutation operations, $M^{i+1}$ is the molecular population generated using $R^{\prime\prime i}$, and $M^i$ is the current molecular population. This selection mechanism ensures that rule updates are retained only when they improve the overall quality of the molecular population.

\subsection{Molecular Evolution Engine}
\label{sec:mol-evo}

The molecular evolution engine manages structural evolution through integrated mutation and crossover operations, working with the rule system to optimize molecules while expanding chemical space exploration.

\begin{figure}[t]
  \centering
  \includegraphics[width=0.9\linewidth]{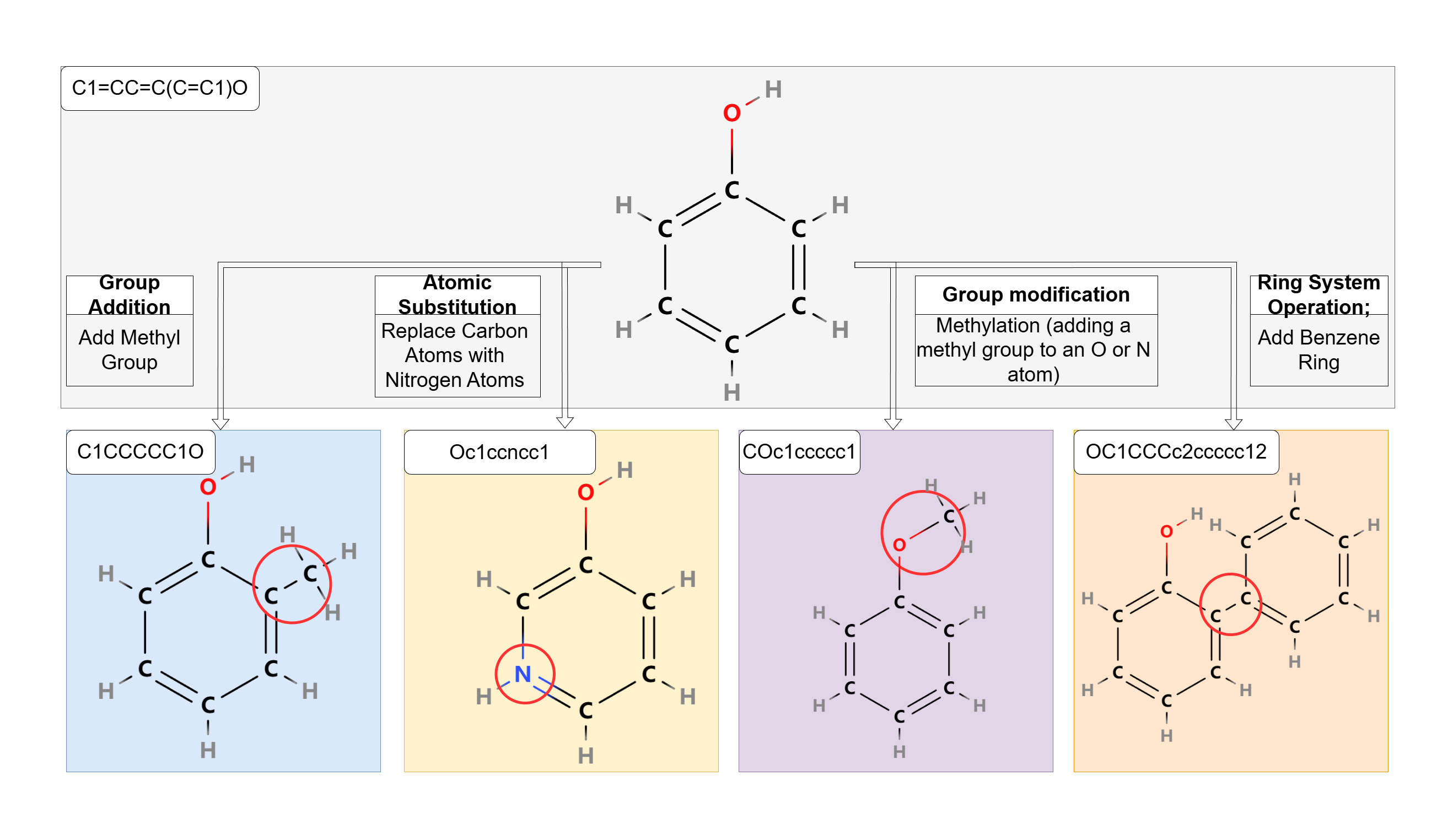}
  \caption{Examples of molecular mutations.}
  \label{bianyi}
\end{figure}

\textbf{Molecular Mutation.}
For mutation operations, we apply the rule-based transformation system defined in Section~\ref{sec3-1}. For each molecule $m \in M$ (where $M$ denotes the molecular population), when the conditions in $C$ are satisfied, the associated transformations $T$ are applied sequentially as defined in Eq.~\eqref{eq:apply}. If any transformation yields null or the conditions are not met, the mutation returns null, ensuring only chemically valid transformations are retained. Fig.~\ref{bianyi} illustrates examples of molecular mutations, showing how different transformation operations modify molecular structures.

\textbf{Molecular Crossover.}
While rule-based mutations ensure interpretability, molecular crossover operations are essential for diversity and innovation. We implement the seven crossover strategies defined in Section~\ref{sec3-3} through an adaptive engine that dynamically selects the most effective strategy based on historical performance. When the selected strategy fails to produce a valid molecule, the engine sequentially attempts alternative strategies. If all strategies fail, the system returns an unmodified parent molecule, ensuring that crossover operations always yield valid outputs.

\begin{figure}[t]
  \centering
  \includegraphics[width=0.9\linewidth]{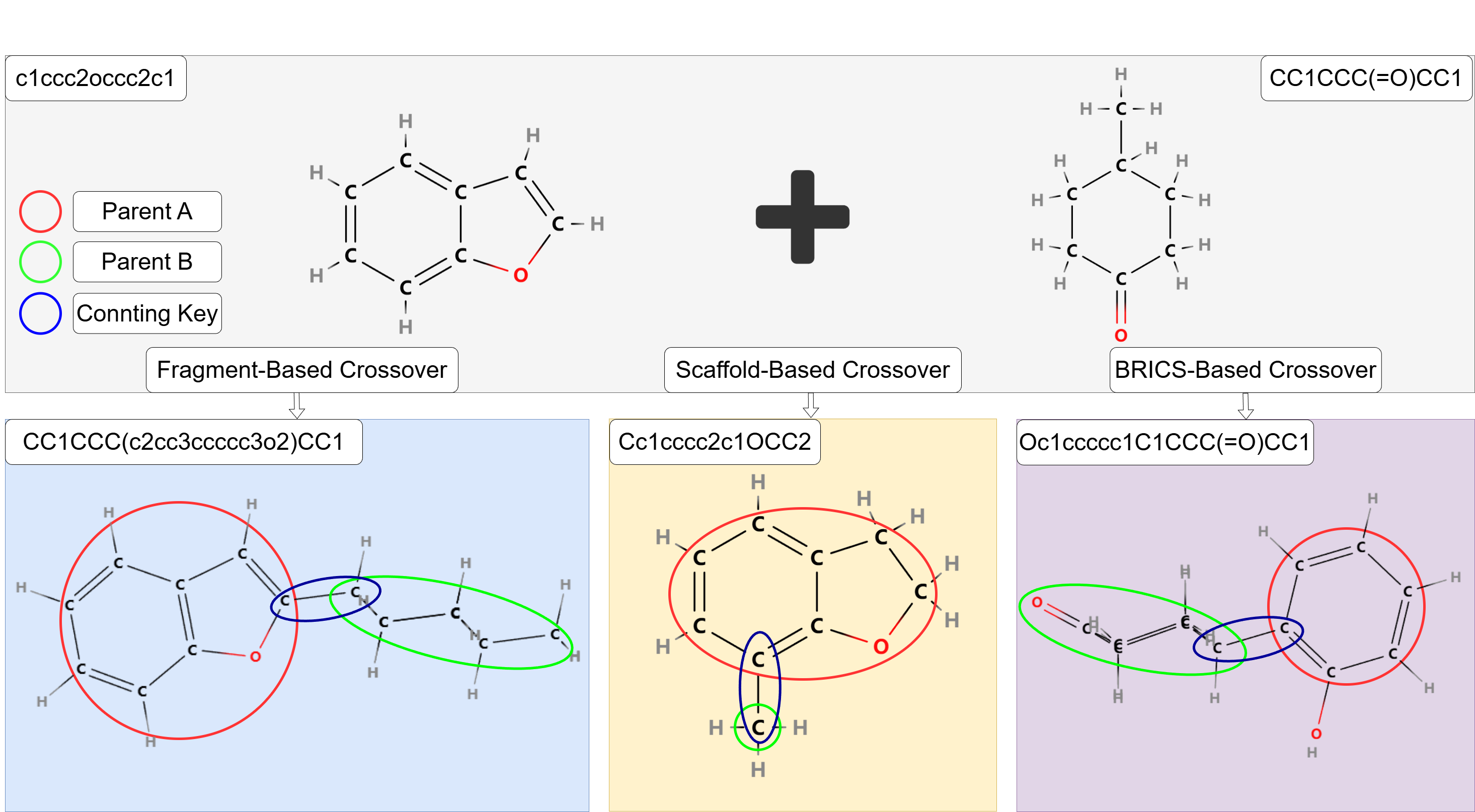}
  \caption{Molecular crossover engine.}
  \label{jiaocha}
\end{figure}

As shown in Fig.~\ref{jiaocha}, using Benzofuran and 4-Methylcyclohexanone as parents A and B, our framework demonstrates three crossover outcomes: fragment-based crossover yields 2-(4-Methylcyclohexyl)benzofuran, scaffold-based crossover produces 7-Methyl-2,3-dihydrobenzofuran, and BRICS crossover generates 2-(4-Oxocyclohexyl)phenol.

To maximize crossover success rate, we employ an adaptive strategy selector that updates the application probabilities of crossover options based on their historical performance, biasing the search toward recently effective strategies \cite{SALGOTRA2024125055}:
\begin{equation}
    \label{eq:xover-prob}
    P_y = \frac{S_y}{\sum_j S_j}
\end{equation}
where $P_y$ is the probability of selecting strategy $y$, and $S_y$ is the success rate of strategy $y$ computed over recent generations. This adaptive mechanism allows the system to learn which crossover strategies work best for the current optimization task and molecular population.

\textbf{Population Management and Update.}
After mutation and crossover operations, each molecule in the population may produce multiple offspring. The engine implements a selection strategy that preserves both diversity and quality:
\begin{equation}
    M_{\text{new}} = \text{top}(\text{unique}(M_{\text{current}} \cup M_{\text{offspring}}), N),
\end{equation}
where $M_{\text{new}}$ is the updated molecular population, $M_{\text{current}}$ the current population, $M_{\text{offspring}}$ the offspring molecules generated through evolutionary operations, and $N$ the maximum population size. The function $\text{unique}(\cdot)$ removes duplicate molecules to maintain diversity, while $\text{top}(\cdot, N)$ performs selection by keeping the top $N$ molecules based on fitness computed using the scoring function from Section~\ref{sec3-2}.

\textbf{Dynamic Parameter Tuning.}
The molecular evolution engine employs population diversity as an adaptive control parameter to balance exploration and exploitation. Calculated as one minus the average pairwise Tanimoto similarity of sampled Morgan fingerprints, this metric is incorporated into rule fitness evaluation to maintain structural diversity, thereby ensuring efficient search and avoiding premature convergence. This adaptive mechanism works in tandem with the rule weight updates described in Section~\ref{sec:rule-evo}, creating a co-evolutionary system where both transformation rules and their application probabilities evolve to optimize molecular design outcomes.

\section{Experimental Setup}
\label{sec:experiments}

We evaluate MCEMOL's performance in molecular generation tasks using standard evolutionary algorithm configurations. The goal is to optimize transformation rules that generate drug-like molecules with high structural diversity, complete validity, and superior drug-likeness properties.

\subsection{Computational Setup}
Experiments were conducted on a laptop with Intel Core i5-12500H CPU and 16GB RAM. MCEMOL completed 100 generations in approximately 45 hours without GPU acceleration, expanding from 20 seed molecules to 829 unique molecules.

\subsection{Dataset}
We use the ZINC250K dataset containing approximately 250,000 drug-like molecules \cite{Irwin2005ZINC}, a standard benchmark that enables direct evaluation of molecular generation methods in drug discovery contexts.

\subsection{Evaluation Framework}

Our evaluation framework combines baseline comparisons with comprehensive drug-likeness assessments to characterize MCEMOL's performance.

\textbf{Baselines and Comparison Data.} We compare against JT-VAE, GCPN \cite{NEURIPS2018_d60678e8,pmlr-v80-jin18a}, GraphVAE, and ORGAN, representing diverse molecular generation approaches. To avoid reimplementation bias, we use results directly from their original papers. Missing metrics are marked as "--" in tables. For MCEMOL, all metrics follow consistent definitions across experiments.

\textbf{Evaluation Metrics and Compliance Criteria.} Molecular structure quality is measured through validity, internal diversity (Tanimoto distance), similarity to references, novelty, and scaffold diversity. The similarity metric uses Morgan fingerprints (ECFP4) with Tanimoto coefficient; while JT-VAE and GCPN track paired molecules during optimization, MCEMOL computes inter-generational similarity between initial and final populations due to its population-based evolution. Property evaluation includes LogP, HOMO-LUMO gap (ATLAS-DMPNN-predicted), QED, molecular weight, synthetic accessibility, TPSA, hydrogen bond acceptors, and rotatable bonds. We assess compliance with Lipinski's Rule of Five, Ghose's filter, and Veber's rule for drug-likeness.

\subsection{Algorithm Configuration}

MCEMOL's configuration balances exploration and exploitation while maintaining chemical validity.

\textbf{Evolutionary Algorithm Parameters.} We use population size 100, 100 generations, rule-level mutation rate 0.4, rule-level crossover rate 0.2, and elitism rate 0.1.

\textbf{Molecular Fitness Evaluation.} Twelve properties are evaluated using normalized weights tailored to optimization objectives. Weights are assigned hierarchically based on impact: Pharmacological relevance (Primary, $0.11\text{--}0.18$) prioritizes Property coverage ($0.18$) and Symmetry ($0.13$); Standard development criteria (Secondary, $0.08$) include Validity, Diversity, Synthetic feasibility, Flexibility, and Drug-like properties; Refinement criteria (Tertiary, $0.05\text{--}0.06$) cover Stability, Pharmacophore, and Drug-likeness; while Structural integrity ($0.02$) serves as a baseline check.

\textbf{Rule Evaluation Criteria.} Rules are evaluated by transformation success rate and resulting molecular fitness scores. Complex conditions and mutations receive higher weights to reflect their contribution to chemical space exploration. These criteria remain fixed across experiments for consistency.

\section{Results}
\label{sec:results}

We divide our experimental analysis into four parts: molecular structure, molecular properties, compliance with drug-like rules, and TOP molecular analysis.

\subsection{Molecular Structure}
We use real-world data from the GCPN and JT-VAE studies~\cite{NEURIPS2018_d60678e8,pmlr-v80-jin18a} to evaluate the performance of MCEMOL across three key metrics.

To avoid reimplementation bias, baseline results are cited directly from original publications, with unreported metrics marked as ``--''. As shown in Table~\ref{t1}, MCEMOL achieves 100\% validity and competitive diversity. Unlike generation-focused baselines, MCEMOL prioritizes discovering interpretable transformation rules via a dual-layer evolutionary mechanism, treating the molecular library as a byproduct rather than the primary objective. Specifically, MCEMOL attains perfect validity, 1.0 novelty (absence from training set), and 0.51 scaffold diversity (unique Murcko scaffolds)—metrics unreported in baseline studies. While its internal diversity (0.834) is comparable to existing methods, MCEMOL achieves the lowest similarity score (0.26), indicating superior exploration of novel chemical space. Furthermore, it uncovers transferable transformation principles, offering both practical molecules and interpretable design knowledge that conventional generative approaches lack.

\begin{table}[htbp]
\centering
\caption{Molecular Structure}\label{t1}
\begin{tabular}{lccccc}
\toprule
Method & Validity (\%) & Diversity & Similarity & Novel & Scaffold Diversity \\
\midrule
JT-VAE & 100 & -- & 0.280 & -- & -- \\
GCPN & 100 & 0.855 & 0.320 & -- & -- \\
GraphVAE & 14 & -- & -- & -- & -- \\
ORGAN & 2.2 & 0.759 & -- & -- & -- \\
MCEMOL & \textbf{100} & \textbf{0.834} & \textbf{0.262} & \textbf{1} & \textbf{0.514} \\
\bottomrule
\end{tabular}
\end{table}

\subsection{Molecular Properties}

\begin{figure}
\centering
\includegraphics[width=0.8\textwidth]{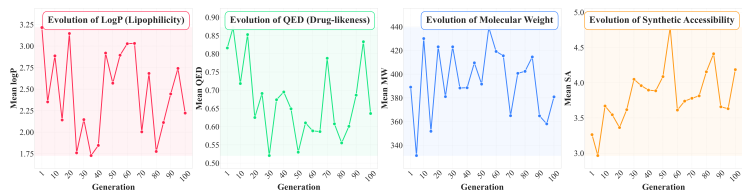}
\caption{Trend diagram of basic chemical properties.}
\label{prop}
\end{figure}

Table~\ref{t2} summarizes average properties, utilizing the ATLAS-DMPNN for efficient HOMO-LUMO gap prediction, while Fig.~\ref{prop} tracks their evolution. LogP decreases within the ideal lipophilic range, and sustained high QED confirms strong drug-likeness. The HOMO-LUMO gap indicates stability and low synthetic difficulty, with molecular weight aligned for oral bioavailability. Synthetic Accessibility (SA) stabilizes at moderate levels, demonstrating controlled complexity. Favorable TPSA, HBA values, and rotatable bond counts ensure permeability and balanced flexibility. These properties collectively show that the generated molecules exhibit excellent drug-like characteristics, making them suitable for pharmaceutical development.

\begin{table}[htbp]
  \centering
  \caption{Molecular Properties}
  \resizebox{\linewidth}{!}{
    \begin{tabular}{cccccccc}
    \hline
    LogP & Gap(eV) & QED & MW(DA) & SA SCORE & TPSA(Å²) & HBA & RotBonds \\
    \hline
    2.41  & 4.4854 & 0.75  & 384.5 & 3.57  & 94.94 & 5.98  & 4.66 \\
    \hline
    \end{tabular}%
    }
  \label{t2}%
\end{table}%

\subsection{Drug-Likeness Compliance}
The molecules generated by this framework exhibit exceptional compliance with drug-likeness rules:

100\% compliance with Lipinski's rule: All molecules meet the fundamental criteria for oral drugs.
98.0\% compliance with Ghose’s rule: The vast majority of molecules align with the physicochemical properties required in medicinal chemistry.
99.6\% compliance with Veber’s rule: Nearly all molecules possess favorable potential for oral bioavailability.
These high compliance rates indicate that the framework can generate drug-like molecules efficiently, significantly reducing the workload for subsequent screening and optimization.
The detailed chemical analysis of top-performing molecules is provided in Section~\ref{sec:results-qual}.

\subsection{Top Molecules}
\label{sec:results-qual}

\vspace{-5pt}
\begin{table}[htbp]
  \centering
  \caption{Top Molecules}
  \resizebox{\linewidth}{!}{
    \begin{tabular}{llll}
    \hline
    \multicolumn{1}{c}{\textbf{SMILES}} & \multicolumn{1}{c}{\textbf{LogP}} & \multicolumn{1}{c}{\textbf{QED}} & \multicolumn{1}{c}{\textbf{Structural Features}} \\
    \hline
    Cc1ccc(F)c(NC(=O)c2ccc(S(=O)(=O)CO)cc2)c1 & 2.11  & 0.902 & Fluorinated aniline structure, sulfonyl side chain. \\
    Cc1cc(C(=O)NC2CC2)c(CO)n1-c1ccc2c(c1)OC=CO2 & 2.41  & 0.905 & Condensation structure of nitrogenous heterocyclic ring. \\
    COc1cc(Cl)c(C)cc1NC(=O)c1ccc(=O)n(CO)c1 & 2.02  & 0.902 & Chloraniline pyridone derivatives. \\
    COCn1cc(C(=O)Nc2nc(C)c(Cl)cc2OC)ccc1=O & 2.07  & 0.903 & Condensation of pyridone and pyridine chloride. \\
    COC(C)c1c(C(=O)NC2CC2)cc(C)n1-c1cc2c(cn1)OCCO2 & 2.55  & 0.89  & Complex heterocyclic system containing dioxane. \\
    \hline
    \end{tabular}%
    }
  \label{t3}%
\end{table}

Analysis of the top-performing molecules (Table~\ref{t3}) demonstrates MCEMOL's ability to generate chemically meaningful structures with diverse pharmacological potential. Taking the first two molecules as examples:

The first molecule (Cc1ccc(F)c(NC(=O)c2ccc(S(=O)(=O)CO)cc2)c1) exhibits distinct drug-like features: a fluorine atom enhances metabolic stability and lipophilicity, while amide and sulfonyl groups act as hydrogen bond acceptors (with the latter improving water solubility). Complemented by a hydroxyl hydrogen bond donor, these traits suggest potential utility as an enzyme inhibitor or anti-inflammatory/analgesic agent.

The second molecule (Cc1cc(C(=O)NC2CC2)c(CO)n1-c1ccc2c(c1)OC=CO2) displays sophisticated structural design. The cyclopropane moiety introduces conformational constraints and metabolic stability, while the dioxane ring may enhance target binding affinity. Supported by multiple hydrogen bond interactions and a versatile hydroxymethyl group, this structure holds promise as a kinase inhibitor or central nervous system (CNS) drug.

These examples illustrate how MCEMOL's dual-layer evolutionary mechanism effectively integrates chemical knowledge with optimization algorithms to yield molecules that are theoretically sound and practically relevant for drug development.

\section{Conclusion}

MCEMOL is a multi-constrained evolutionary molecular design framework that integrates rule evolution with direct molecular manipulation~\cite{https://doi.org/10.1002/minf.201700123}. By combining explainable rule-based transformations with evolutionary optimization, it addresses key challenges in AI-driven molecular design while maintaining interpretability. 

Our results demonstrate five main achievements. First, dual-layer evolution at both rule and molecule levels outperforms existing methods through co-evolutionary exploration of chemical space~\cite{https://doi.org/10.1002/minf.201700123,Olivecrona2017}. Second, the multi-dimensional constraint system ensures molecular quality in synthesis feasibility, symmetry, and pharmacophore representation~\cite{Polishchuk2013,Hoksza2014}. Third, adaptive rule weighting and dynamic pool management enhance learning efficiency and adaptation to diverse design goals~\cite{Liu2017,Hansen2006}. Fourth, the framework achieves high interpretability without sacrificing performance~\cite{D1SC05259D}. Fifth, MCEMOL's computational efficiency enables practical deployment from minimal starting molecules, making it accessible for resource-limited research settings.

Experiments show strong novelty, diversity, and drug-likeness with 100\% validity and near-perfect pharmaceutical rule compliance while retaining interpretability~\cite{Jimenez-Luna2020,D1SC05259D}. The qualitative analysis in Section~\ref{sec:results-qual} confirms chemical plausibility and pharmacological potential.

Current limitations include dependence on predefined rule libraries that may restrict exploration of unconventional chemical space, challenges in handling flexible 3D conformations~\cite{Arus-Pous2019}, and the need for hyperparameter tuning for different objectives.

Future work will explore Pareto and multi-objective Bayesian strategies, integrate synthesis planning, incorporate protein structures and bioactivity data for target specificity, investigate fragment-based design, and add medicinal chemistry expert rules.

Overall, MCEMOL provides an interpretable and efficient molecular design approach delivering mechanistic insights through transformation rules and a diverse library of high-quality drug-like molecules, helping bridge computational design and practical drug discovery.

%
%
%
%
\bibliographystyle{splncs04}
\bibliography{reference}
\end{document}